\documentclass[10pt,twocolumn,letterpaper]{article}

\usepackage{wacv}              

\definecolor{wacvblue}{rgb}{0.21,0.49,0.74}
\usepackage[pagebackref,breaklinks,colorlinks,allcolors=wacvblue]{hyperref}

\def\wacvPaperID{*****} 
\def\confName{WACV}
\def\confYear{2026}

\title{From Prompts to Deployment: Auto-Curated Domain-Specific Dataset Generation via Diffusion Models}

\author{Dongsik Yoon\\
HDC LABS\\
Seoul, Republic of Korea\\
{\tt\small kevinds1106@hdc-labs.com}
\and
Jongeun Kim\\
HDC LABS\\
Seoul, Republic of Korea\\
{\tt\small JongeunKim@hdc-labs.com}
}

\begin{document}
\maketitle
\begin{abstract}
In this paper, we present an automated pipeline for generating domain-specific synthetic datasets with diffusion models, addressing the distribution shift between pre-trained models and real-world deployment environments. Our three-stage framework first synthesizes target objects within domain-specific backgrounds through controlled inpainting. The generated outputs are then validated via a multi-modal assessment that integrates object detection, aesthetic scoring, and vision–language alignment. Finally, a user-preference classifier is employed to capture subjective selection criteria. This pipeline enables the efficient construction of high-quality, deployable datasets while reducing reliance on extensive real-world data collection.

\end{abstract}

\section{Introduction}
\label{sec:intro}
With the rapid progress of deep learning, large-scale datasets and their corresponding pre-trained models have become widely accessible. Nevertheless, models pre-trained on public corpora often perform poorly on real-world deployments due to distribution shift between the source (training) data and the target environment. Public benchmarks frequently inadequately cover domain-specific factors such as particular geographic locations, uncommon viewpoints, or specialized object categories—thereby limiting the immediate adaptability of pre-trained models.
Moreover, acquiring datasets that capture specific sites, atypical camera viewpoints, or specialized object categories is substantially more demanding than building general-purpose datasets, typically incurring higher collection, curation, and annotation costs.

\begin{figure*}[!t]
  \includegraphics[scale=0.4]{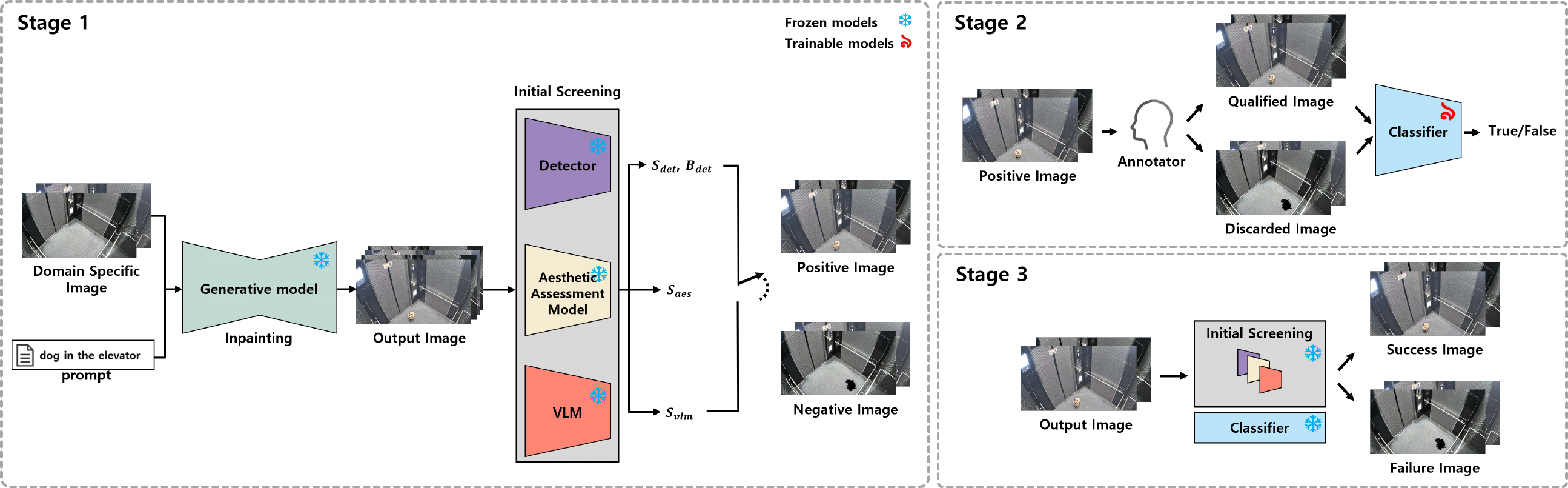}
  \caption{Overview of the proposed three-stage, diffusion-based dataset generation and auto-curation pipeline.}
  \label{fig:fig1}
  \vspace{-3mm}
\end{figure*}

For instance, training a robust pet detection model for elevator surveillance systems requires extensive footage of animals captured from overhead CCTV angles—a viewpoint rarely represented in standard object detection benchmarks such as COCO~\cite{coco} or ImageNet~\cite{imagenet}. Similarly, tasks such as detecting fires in underground parking structures or recognizing robot vacuum cleaners in home-security footage require highly specialized datasets that are expensive to gather and often unsafe to stage in real environments.

Prior work has addressed these limitations through three complementary approaches: (1) domain adaptation techniques~\cite{adapt1, adapt2} that align feature distributions or learn domain-invariant representations~\cite{domain1} to reduce source-target distribution shifts (2) few-shot fine-tuning methods~\cite{fewshot} that adapt pre-trained models using limited labeled target examples and (3) diffusion-based generative models~\cite{synthetic} that synthesize target-aligned datasets and augment existing corpora with rare scenes~\cite{effective}, viewpoints, and object categories. Recent progress in diffusion models~\cite{ddpm} has markedly lowered the cost of constructing large-scale synthetic datasets, and such data are now used to strengthen public benchmarks.

However, naively applying diffusion models to dataset generation introduces its own challenges. Generated images frequently suffer from prompt misalignment, where synthesized objects deviate from the intended specifications in terms of category, pose, or scale. Additionally, aesthetic artifacts and unrealistic object placements can degrade downstream model performance if such images are included in training sets without proper filtering. These issues necessitate a systematic curation process that goes beyond simply adding generated images.

Building on advances in diffusion-based generative modeling, we aim to generate images of difficult-to-capture viewpoints and object categories and to design a multi-stage curation pipeline that converts these synthetic outputs into high-quality training datasets. Our pipeline first generates candidate images from user-specified viewpoints and target objects using a diffusion-based generator, then validates each image through three sequential filtering stages. First, to address the chronic issue in diffusion models where synthesized objects often deviate from intended prompts, we assess text-image alignment by comparing VLM-generated captions against input prompts. Second, we filter low-quality outputs using aesthetic scoring and object detection. Finally, we deploy a DINOv2-based classifier to eliminate residual failures such as awkward poses and sizes that persist despite prior filtering.
Our automated pipeline thus yields high-quality synthetic datasets, improving deployment reliability in target environments without the burden of extensive real-world data collection.

Our contributions are twofold: (1) we propose an end-to-end pipeline that integrates diffusion-based synthesis with multi-modal quality assessment for domain-specific dataset generation, and (2) we introduce a simple yet effective preference classifier that learns user-specific curation criteria from minimal annotations, enabling scalable and personalized dataset construction.
\section{Method}
\label{sec:method}

Our automated dataset generation pipeline has two prerequisites:
\begin{enumerate}
    \item \textbf{Domain-specific background images}: scenes that are intrinsically difficult to collect or tied to a specific site (e.g., underground parking lots, elevator CCTV views, and home security camera footage). Because these images constitute the backgrounds in which the model will operate in real-world settings, we recognize that the number of obtainable images may be limited.
    \item \textbf{Target object}: the object to be synthesized within the given domain context. While these objects are typically common and could feasibly appear in domain images, we focus on natural object--scene combinations that are scarce in publicly available datasets (e.g., underground parking lot with a fire, elevator CCTV viewpoint with a dog, robot vacuum cleaners in home security footage).
\end{enumerate}

\subsection{Stage 1: Object Synthesis and Initial Screening}

With these two key factors, Stage 1 inserts the target object into domain images using state-of-the-art image generators (e.g., Stable Diffusion~\cite{stablediffusion}, Midjourney and FLUX~\cite{flux}).
We then define a prompt \(P\) that explicitly requires the presence of the target object within the specified domain scene. Guided by \(P\), the framework performs inpainting within a user-defined region of interest (ROI), where random mask locations are sampled. The mask size corresponds to the expected size of the object as specified by the user.

To evaluate Stage-1 outputs, we apply a three-component protocol: (i) an object detector~\cite{detector}, (ii) an aesthetics assessment model~\cite{aesthetic}, and (iii) a vision--language model (VLM)~\cite{vlm}. For each inpainted image, the detector verifies category-level presence of the object, the aesthetics model filters out low-quality generations, and the VLM evaluates prompt--image alignment to confirm the inclusion of the target object.

For the detector, we extract the target-class confidence \(S_{\mathrm{det}}\) and the predicted bounding box \(B_{\mathrm{det}}\).
Then, to verify spatial fidelity, we compute IoU between \(B_{\mathrm{det}}\) and the inpainting mask region \(M\).
A high IoU indicates that the synthesized object is contained within the designated region and that its extent approximately matches the intended size.
To further evaluate the outputs, we incorporate an aesthetics assessment model to measure the overall visual quality and completeness of the generated images. The aesthetic score \(S_{\mathrm{aes}}\) complements the detector’s confidence score by providing an additional indicator of image fidelity and perceptual quality.
We further employ a VLM to evaluate prompt-image alignment. Given both the generated image and \(P\), the VLM produces a descriptive caption \(C\) of the entire image. We then compute the cosine similarity $S_{\mathrm{vlm}} = \cos\!\big(\phi(C),\, \phi(P)\big)$ using sentence embeddings \(\phi(\cdot)\)~\cite{sentence}.
By integrating the detector, aesthetics model, and VLM, we evaluate whether the synthesized images faithfully represent the user-specified target objects within the intended domain.

In our case study (dog-in-elevator scenario), we use the following thresholds: detector confidence $S_{det} > 0.8$, aesthetics score $S_{aes} > 5$, spatial fidelity $IoU(B_{det}, M) > 0.8$, and VLM alignment $S_{vlm} > 0.8$. These thresholds are adjustable and may vary with the choice of pre-trained models. Figure 2 illustrates failure cases from Stage 1: images with low $S_{det}$ and $S_{aes}$ scores that fail to synthesize the target object consistently. Even when both $S_{det}$ and $S_{aes}$ thresholds are met, the VLM-based $S_{vlm}$ metric offers a complementary validation layer drawn from a different model family to suppress anomalous outputs.

  \begin{figure}[!t]
    \centering
    \includegraphics[scale=0.7]{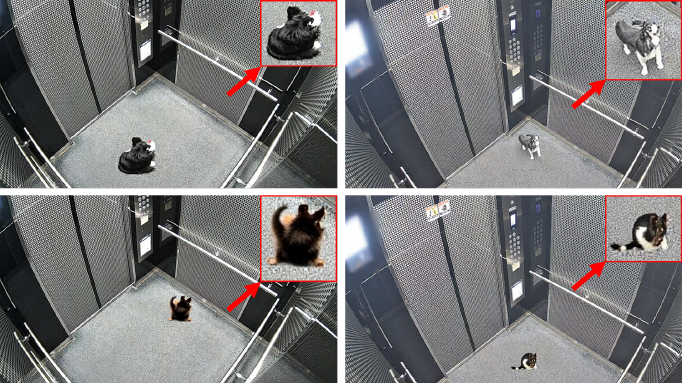}
    \caption{Negative images that failed object synthesis in Stage 1 due to low detection and aesthetics scores.}
    \label{fig:fig2}
  \end{figure}

  \begin{figure}[!t]
    \centering
    \includegraphics[scale=0.7]{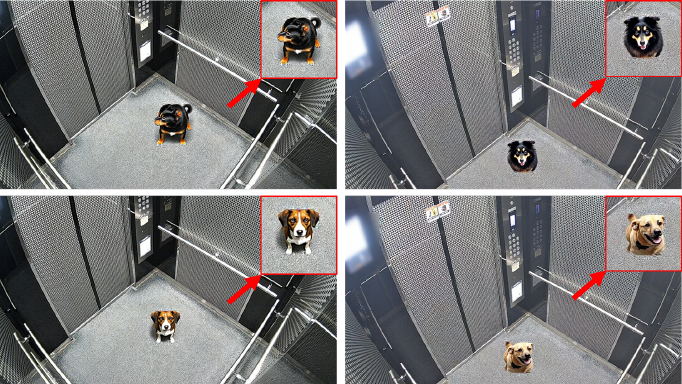}
    \caption{Discarded images rejected in Stage 2 due to poor viewpoint/pose from annotators despite successful object synthesis.}
    \label{fig:fig3}
  \end{figure}

\subsection{Stage 2: Preference Classification}
In the second stage, images deemed successful in Stage 1 are subjected to an additional round of user evaluation. While effective at removing flawed generations, deep learning based filtering models cannot accommodate user-specific preferences, such as the desired pose or viewing angle of the target object within the scene. To address this limitation, we incorporate a preference classifier: users manually annotate generated images as either successful or failed, and these annotations are then used to train a classifier for automatic filtering. This component enables the pipeline not only to eliminate flawed outputs but also to account for user preferences.


Our hybrid architecture (DINOv2+ConvNeXt) combines DINOv2-base (86M parameters) for global semantic reasoning with ConvNeXt-base~\cite{convnet} (89M parameters) for fine-grained spatial analysis. We initialize freezing ratios of 0.8 for the DINOv2 encoder and 0.7 for the ConvNeXt stages. For both backbones, inputs are first restricted to the region of interest through an expand-and-crop step (expand ratio 0.3). The resulting features are then refined by ROI attention modules, which are lightweight two-layer MLPs with ReLU activation and a sigmoid gate that provide content-adaptive weighting within the cropped ROI. The attended features are fused through multi-scale processing, and the fused representation is mapped to the output via a three-layer MLP head. The total parameter count is approximately 176M.

Figure 3 presents samples rejected by annotators from stage2. Although the target object is synthesized, these images are either poorly integrated with the domain-specific viewpoint or deviate from the desired viewpoint/pose. We train the DINOv2-based preference classifier on both accepted and rejected samples to capture user preferences, enabling subsequent generations to be filtered or re-ranked accordingly.

\subsection{Stage 3: Final Dataset Construction}
Finally, in the third stage, we integrate the models from Stage 1 with the preference classifier from Stage 2 to construct a high-quality dataset that reflects both objective quality measures and user intent.
Figure 4 shows successful images that pass all stages. These results integrate harmoniously with the user-provided, domain-specific background and depict the target object with the desired viewpoint, pose, and scale. This stage requires no additional annotation, as the classifier generalizes from the pilot set collected in Stage 2. Overall, the pipeline facilitates construction of object datasets that are otherwise difficult to obtain in specialized domain settings.
\section{Experiments}

  \begin{figure*}[!t]
    \centering
    \includegraphics[scale=1]{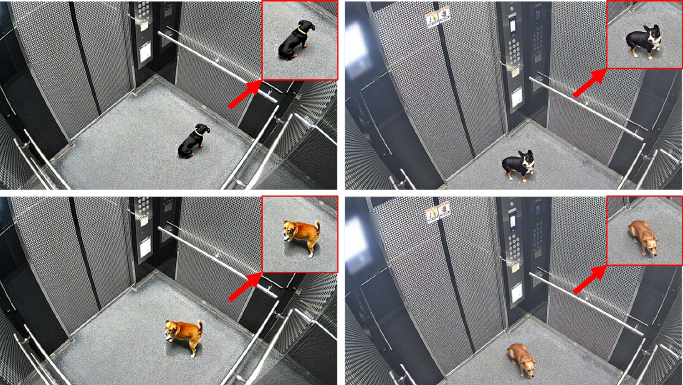}
    \caption{Success image that passes all three stages, demonstrating target object synthesis within a domain-specific background.}
    \label{fig:fig4}
  \end{figure*}  

\begin{table}[]
\begin{tabular}{|l||c|c|c|}
\hline
Backbone Architecture       & Precision & Recall & F1-score \\ \hline\hline
VGG19 & 0.7733         & 0.8940      & 0.8293        \\ \hline
ViT-Large/16   & 0.7492         & 0.9152      & 0.8239        \\ \hline
CLIP (ViT-Large/14)  & 0.7575         & \textbf{0.9210}      & 0.8313        \\ \hline
Ours  & \textbf{0.7822}        & 0.9133     & \textbf{0.8427}       \\ \hline
\end{tabular}
\caption{Backbone comparison for preference classification: best in bold.}
\vspace{-3mm}
\end{table}

In our experiments, we use elevator CCTV views as the domain images and dogs as the target object class. To demonstrate the effectiveness of our proposed preference classifier, we compare it against various existing CNN backbone classifiers. For this evaluation, we utilize images that successfully passed Stage 1 filtering. Five annotators reviewed these images and labeled them based on subjective preference criteria, resulting in a training dataset of approximately 300 images and a test dataset of approximately 1,000 images. 

\subsection{Implementation Details}
All experiments are implemented in PyTorch. We train four different backbone architectures on a curated dataset of 289 images (166 positive, 123 negative samples) with an 80:20 train-validation split. All models are trained for 50 epochs with early stopping patience of 10 epochs based on validation F1-score. We employ AdamW optimizer with learning rate 2e-5, weight decay 1e-4, and cosine annealing scheduler with 10\% warmup ratio. Gradient clipping with maximum norm 1.0 is applied for training stability. All models are evaluated on the same test set of 991 images (519 positive, 472 negative) using precision, recall, and F1-score as primary metrics.

\subsection{Backbone Details}
 For the backbone architecture comparison, we employ VGG-19~\cite{vgg19}, ViT-Large/16~\cite{vit}, and CLIP (ViT-Large/14)~\cite{clip}, all initialized with pretrained weights and trained with a backbone freezing ratio of 0.8. As in our proposed model, inputs are first expand-and-cropped around the ROI, and the extracted features are refined by ROI-attention modules. The attended features are then passed to a three-layer MLP classifier with nonlinear activations (ReLU for VGG; GELU for ViT and CLIP) and a dropout rate of 0.3. The parameter counts are 190M (VGG-19), 304M (ViT-Large/16), and 305M (CLIP).
 

As shown in Table 1, our proposed DINOv2-based~\cite{dinov2} classifier outperforms all baseline architectures in F1-score despite its conceptual simplicity. Notably, our approach requires only approximately 300 annotated images to learn effective preference boundaries—a fraction of what traditional supervised learning typically demands. This result suggests that pretrained representations, combined with lightweight adaptation, can effectively capture domain-specific preferences without requiring complex architectures or large-scale annotations.
\section{Conclusion}
We introduced a multi-stage diffusion-based pipeline for domain-specific dataset generation and automated curation. The framework integrates an object detector, an aesthetics assessor, a vision–language model for prompt–image alignment, and a trained preference classifier to incorporate user-specific criteria. By combining these components, our approach reduces the cost and complexity of acquiring domain-specific training data while ensuring generated dataset quality through systematic curation.

While promising, our study has several limitations. First, we evaluate the pipeline only on a single domain–object pair (elevator CCTV with dogs), so cross-domain generalization remains to be demonstrated. Second, the pipeline assumes access to a modest set of domain-specific background images; in extremely data-scarce settings, collecting such seed images may still be a bottleneck even though our approach reduces the need for large-scale, risky real-world data collection. Third, the filtering stages rely on off-the-shelf pre-trained detectors, aesthetics models, and VLMs, and we leave both a more systematic analysis of their biases and failure modes and a full ablation of all three stages to future work. 

In future work, we also plan to automate threshold selection in Stage 1, incorporate iterative user feedback for continuous classifier improvement, and investigate the impact of synthetic data scale on downstream task performance.

\clearpage
{
    \small
    \bibliographystyle{ieeenat_fullname}
    \bibliography{main}
}

\end{document}